\documentclass[10pt,twocolumn,letterpaper]{article}

\usepackage{wacv}              

\usepackage{graphicx}
\usepackage{amsmath}
\usepackage{amssymb}
\usepackage{booktabs}
\usepackage[accsupp]{axessibility}  

\usepackage{enumitem,amssymb}
\usepackage{todonotes}
\usepackage{pifont}

\usepackage{url}
\usepackage{color, colortbl}
\usepackage{booktabs}
\usepackage{fixltx2e}
\usepackage{floatrow}  
\usepackage{blindtext}  
\usepackage{array, caption, floatrow, makecell, booktabs}  
\usepackage{multirow}  
\usepackage{rotating}  
\usepackage{color,soul}
\usepackage{amsmath}
\usepackage{mathtools}

\usepackage{fdsymbol}

\usepackage{amsmath}
\usepackage{dsfont}
\usepackage{mathtools}

\usepackage{algorithmic}

\usepackage{amsmath}
\usepackage{xparse}

\ExplSyntaxOn
\NewDocumentCommand{\ucgreek}{m}
 {\str_case:nn { #1 } {
    {A}{\mathrm{A}} {B}{\mathrm{B}} {C}{\Sigma} {D}{\Delta} {E}{\mathrm{E}} 
    {F}{\Phi} {G}{\Gamma} {H}{\mathrm{H}} {I}{\mathrm{I}} {J}{\Theta} {K}{\mathrm{K}} 
    {L}{\Lambda} {M}{\mathrm{M}} {N}{\mathrm{N}} {O}{\mathrm{O}} {P}{\Pi}
    {Q}{\mathrm{X}} {R}{\mathrm{P}} {S}{\Sigma} {T}{\mathrm{T}} {U}{\Upsilon} 
    {W}{\Omega} {X}{\Xi} {Y}{\Psi} {Z}{\mathrm{Z}}
}}
\NewDocumentCommand{\lcgreek}{m}
 {\str_case:nn { #1 }
   {{a}{\alpha} {b}{\beta} {c}{\varsigma} {d}{\delta} 
    {e}{\varepsilon} {f}{\varphi} {g}{\gamma} {h}{\eta} {i}{\iota}
    {j}{\vartheta} {k}{\kappa} {l}{\lambda} {m}{\mu} {n}{\nu} {o}{o}
    {p}{\pi} {q}{\chi} {r}{\rho} {s}{\sigma} {t}{\tau} {u}{\upsilon} 
    {w}{\omega} {x}{\xi} {y}{\psi} {z}{\zeta}
}}
\ExplSyntaxOff

\newcommand{\defeq}{\vcentcolon=}
\def\code#1{\texttt{#1}}

\usepackage{pifont}

\usepackage[normalem]{ulem}

\definecolor{lightgray}{rgb}{0.92,0.92,0.92}
\newcolumntype{g}{>{\columncolor{lightgray}}c}
\newcolumntype{?}{!{\vrule width 1.5pt}}  

\newfloatcommand{capbtabbox}{table}[][\FBwidth]

\newlist{todolist}{itemize}{2}
\setlist[todolist]{label=$\square$}

\graphicspath{{figures/}}

\usepackage[pagebackref,breaklinks,colorlinks]{hyperref}

\usepackage[capitalize]{cleveref}
\crefname{section}{Sec.}{Secs.}
\Crefname{section}{Section}{Sections}
\Crefname{table}{Table}{Tables}
\crefname{table}{Tab.}{Tabs.}

\def\wacvPaperID{2007} 
\def\confName{WACV}
\def\confYear{2024}

\begin{document}

\title{Controllable Image Synthesis of Industrial Data using Stable Diffusion}

\author{
    Gabriele Valvano\thanks{Equal Contribution} \\
    Baker Hughes, Florence, Italy \\
    {\tt\small gabriele.valvano[at]bakerhughes.com}
    \and
    Antonino Agostino* \\
    Baker Hughes, Florence, Italy \\
    {\tt\small antonino.agostino[at]bakerhughes.com}
    \and
    Giovanni De Magistris\\
    Baker Hughes, Florence, Italy \\
    \and
    Antonino Graziano \\
    Baker Hughes, Florence, Italy \\
    \and
    Giacomo Veneri \\
    Baker Hughes, Florence, Italy \\
}
\maketitle

\begin{abstract}
    Training supervised deep neural networks that perform defect detection and segmentation requires large-scale fully-annotated datasets, which can be hard or even impossible to obtain in industrial environments. 
    Generative AI offers opportunities to enlarge small industrial datasets artificially, thus enabling the usage of state-of-the-art supervised approaches in the industry. 
    Unfortunately, also good generative models need a lot of data to train, while industrial datasets are often tiny. Here, we propose a new approach for reusing general-purpose pre-trained generative models on industrial data, ultimately allowing the generation of self-labelled defective images. 
    First, we let the model learn the new concept, entailing the novel data distribution. 
    Then, we force it to learn to condition the generative process, producing industrial images that satisfy well-defined topological characteristics and show defects with a given geometry and location. 
    To highlight the advantage of our approach, we use the synthetic dataset to optimise a crack segmentor for a real industrial use case. 
    When the available data is small, we observe considerable performance increase under several metrics, showing the method's potential in production environments.\let\thefootnote\relax\footnotetext{
        Published as: \textit{Proceedings of the IEEE/CVF Winter Conference on Applications of Computer Vision (WACV), 2024, pp. 5354-5363}
    }
\end{abstract}

\section{Introduction}
\label{sec:introduction}

\begin{figure}
    \centering
    \includegraphics[width=0.85\columnwidth]{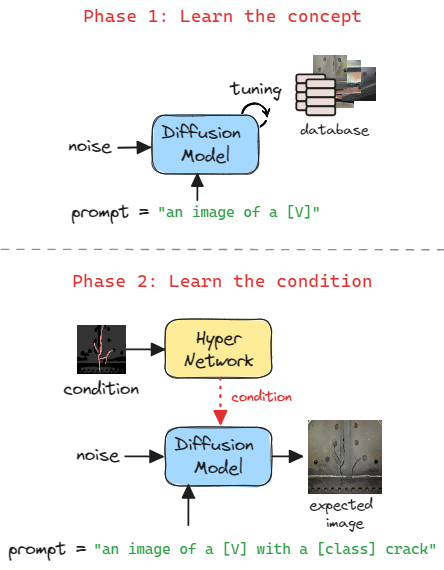}
    \caption{
        Method overview. 
        Our approach involves two main steps: i) {\color[HTML]{e03131}\textit{learn the concept}} and ii) {\color[HTML]{e03131}\textit{learn the condition}}. For the first step, we utilise a high-capacity model that has previously learned a general-purpose image prior. For simplicity, we use Stable Diffusion, which we inject with the knowledge of a new, previously unseen concept. 
        In the second phase, we enforce label-driven constraints to ensure the generation process adherence to specific criteria. 
        Finally, with the resulting conditional generator, we can produce self-annotated data that we can leverage to optimise other models to perform supervised downstream tasks, such as crack detection and segmentation.
    }
    \label{fig:graphical_abstract}
\end{figure}

Over the past decades, artificial intelligence has made significant advancements, leading to several revolutionising industrial applications \cite{zhou2021review,rombach2022high,touvron2023llama}. 
Due to its simplicity, supervised training is one of the most popular approaches to developing vision models for industrial object detection and segmentation. Unfortunately, supervised training of vision models requires large-scale fully-annotated datasets, which can be hard to obtain. 
For instance, in medical imaging, it is often challenging to create datasets containing rare pathologies. Similarly, in production pipelines, collecting datasets containing defective images can be problematic because engineers strive to optimise real-world industrial procedures and minimise the number of defective products as much as possible. Lastly, producing high quality annotations is costly and often requires experts.
To make progress in democratising and incorporating AI into industrial applications, it is essential to minimise the data and annotation requirements needed for training supervised models.

While detecting defects with zero-shot and few-shot learning techniques is an active field of study \cite{huang2022survey,song2023comprehensive}, these techniques still do not adequately address the problem of data scarcity in real-world scenarios. 
In fact, they train models using only small amounts of annotated samples while leaving them ignorant about the entire data distribution, which can be complex and have many unseen modes. 
On the other hand, approaches aiming to reduce the amount of supervision while exploiting larger datasets constitute an exciting perspective \cite{cheplygina2019not,chen2022semi}. In fact, a relatively small amount of unlabelled data is usually much easier to obtain. For example, collecting defect-free images is less complicated than collecting defective ones in industrial procedures, where defects are rare.

Along with this line of thought, we propose using generative models to augment the dataset size artificially. In particular: \textbf{i)} we exploit a generic image prior learned by a pre-trained high-capacity model, Stable Diffusion \cite{rombach2022high}; \textbf{ii)} then, we inject new knowledge about the concept of interest; and finally, \textbf{iii)} we force the generation process to adhere to specific label-driven constraints. 
We provide a graphical overview of the approach in \Cref{fig:graphical_abstract}. Once the generative model is ready, it is possible to use it to generate self-annotated data that we can use to optimise an instance segmentation model and perform crack detection and segmentation.
To summarise, our \textbf{contributions} are:
\begin{itemize}
    \item We design a data generation pipeline for industrial use cases, where images fall out of the distribution used to train general-purpose generative models. The pipeline consists in two main steps: i) \textit{learn the concept}, and ii) \textit{learn the condition}.
    
    \item We introduce a novel conditioning mechanism providing coarse, unsupervised geometrical cues. With this condition, we drive the generative process to produce data with a given geometry and showing defects in the required position. 
    The crucial advantage of these generated data is having annotation masks that are known by construction.
    
    \item We show the advantage of our approach in a real industrial use case. First, we generate new self-annotated data. Then, we use these data to optimise a downstream supervised model and perform instance segmentation. To the best of our knowledge, this is the first work demonstrating the advantage of synthetic data in instance segmentation of industrial combustors.
\end{itemize}

The remainder of the paper is organised as follows. \Cref{sec:related_work} briefly summarises related literature, while \Cref{sec:background} introduces the technical background needed for the subsequent sections. \Cref{sec:method} provides technical details, design choices and methods used by our approach, while \Cref{sec:experiments} reports experimental results and discussions. Finally, \Cref{sec:conclusion} concludes the manuscript.

\section{Related Work}\label{sec:related_work}
\paragraph{Learning with Limited Data}
Having access to large-scale annotated datasets is crucial for training accurate and robust computer vision models. However, collecting and annotating large amounts of data is not always possible, especially for tasks that require fine-grained annotations or experts. In the literature, several techniques aim to reduce data requirements.
Data augmentation involves applying various transformations to existing data to increase its diversity and quantity, which are needed to train better models. 
In this context, generative models offer opportunities to surpass simple data augmentation techniques, synthesising non-trivial but still realistic data variations conditioned on input classes, text, or given geometrical constraints \cite{antoniou2017data,brock2018large,zhang2019self,rombach2022high,wang2022semantic,zhang2023adding,shivashankar2023semantic}. 
Yet, training good conditional generative models from scratch may also require large-scale annotated datasets\cite{park2019semantic,lorenz2019unsupervised,wang2022semantic},\footnote{Note that annotations needed by generative models are often easier to obtain w.r.t. what is usually required by other computer vision tasks, such as instance segmentation. Moreover, some conditional generative models do not need training annotations at all. In this work, we focus on mask-guided image synthesis that is useful for downstream object detection and instance segmentation tasks.} 
leaving their applicability in industrial environments confined to limited settings.
On the other hand, recent years witnessed the open-sourcing of powerful generative models pre-trained on large-scale datasets \cite{rombach2022high,saharia2022photorealistic} and able to provide high-quality results, impossible to distinguish from the real data. The image prior learned by these models serves as a powerful knowledge base that can be exploited in a wide range of downstream tasks, from editing \cite{mou2023dragondiffusion} to inpainting \cite{avrahami2022blended,levin2023differential,lugmayr2022repaint}, keypoint correspondence \cite{peebles2022gan,mu2022coordgan,tang2023emergent}, and few-shot learning \cite{hu2021lora,gal2022image,ruiz2023dreambooth,ruiz2023hyperdreambooth}.
Along with the existing perspectives of foundation models, we propose to address the problem of limited data sources by adapting a large pre-trained model in two main steps: i) learning new concepts and ii) learning to condition concept-based generation. 

\paragraph{Diffusion Models for Conditional Image Synthesis}
Denoising Diffusion Models (DDMs) are generative models trained to learn the underlying data distribution by gradually transforming a noise-corrupted version of the input data towards the original noise-free data~\cite{yang2022diffusion,croitoru2023diffusion,ho2020denoising,nichol2021glide}. 
DDMs use an autoregressive approach consisting in a sequence of denoising steps modelling conditional distributions given previous denoised samples. 
This is made possible by sampling from a defined reverse (denoising) diffusion process that reduces the noise level over time. 
DDMs have demonstrated impressive results in tasks like image generation~\cite{dhariwal2021diffusion,rombach2022high} and image-to-image translation~\cite{meng2021sdedit,saharia2022palette,wang2022semantic,zhang2023adding}, capturing complex data distributions and generating high-quality samples. 
We provide mathematical details about this class of generative models in \Cref{sec:background}.

\paragraph{Image Inpainting}
Image inpainting aims at filling in missing or corrupted regions of an image with plausible content, often by using surrounding information as a reference~\cite{saharia2022palette,avrahami2022blended,elharrouss2020image}. 
Although we could frame our goal as an inpainting task where we generate synthetic defects on top of defect-free images, our method is more general. Indeed, we generate entire defective images, using only weak topological conditioning rather than providing an actual image as input context. 
Thus, the variety of synthetic data our method can produce is much larger than what is attainable by simple inpainting. 
Despite comparing with inpainting methods is not our scope, we conducted a simple experiment measuring data variety obtained by inpainting vs our approach. Our method achieves better mode coverage and data generation quality, as illustrated in \Cref{subsec:our_vs_inpaint}.

\section{Technical Background}\label{sec:background}

\paragraph{Diffusion Models.}
As briefly introduced in \Cref{sec:related_work}, DDMs~\cite{yang2022diffusion,croitoru2023diffusion,ho2020denoising,nichol2021glide,nichol2021improved} are probabilistic models that learn to generate data by denoising normally distributed variables. 
To carry out DDMs training, a forward diffusion process adds Gaussian noise on an image $x_0$ over time, resulting in a fixed Markov Chain of length $T$. 
During this process, we optimise the model to invert the diffusion process and reconstruct uncorrupted samples starting from pure noise (a.k.a. reverse diffusion process).

Given a sample $x_0\!\sim\!q(x)$ (where $q(x)$ is the real data distribution), the forward diffusion process adds a small amount of Gaussian noise to $x_0$, at each step, according to a variance schedule $\{\beta_t \in (0, 1)\}_{t=1}^T$.
We can formalise it as:
\begin{equation}\label{eq:ddm_forward}
    q(\mathbf{x}_t \vert \mathbf{x}_{t-1}) = \mathcal{N}(\mathbf{x}_t; \sqrt{1 - \beta_t} \mathbf{x}_{t-1}, \beta_t\mathbf{I})
\end{equation}
\begin{equation}\label{eq:ddm_forward_cum}
    \quad
    q(\mathbf{x}_{1:T} \vert \mathbf{x}_0) = \prod^T_{t=1} q(\mathbf{x}_t \vert \mathbf{x}_{t-1})
\end{equation}

Note that if $T$ is large enough, $x_T$ is equivalent to an isotropic Gaussian distribution.

During the reverse diffusion process, the model attempts to revert the forward noising process, mapping input noise samples onto realistic noise-free images. 
Formally:
\begin{equation}\label{eq:ddm_reverse}
    p_\theta(\mathbf{x}_{0:T}) = p(\mathbf{x}_T) \prod^T_{t=1} p_\theta(\mathbf{x}_{t-1} \vert \mathbf{x}_t)
\end{equation}
\begin{equation}\label{eq:ddm_reverse_cum}
    \quad
    p_\theta(\mathbf{x}_{t-1} \vert \mathbf{x}_t) = \mathcal{N}(\mathbf{x}_{t-1}; \boldsymbol{\mu}_\theta(\mathbf{x}_t, t), \boldsymbol{\Sigma}_\theta(\mathbf{x}_t, t))
\end{equation}

For simplicity, people often assume $\boldsymbol{\Sigma}_\theta(\mathbf{x}_t, t) = \sigma^2_t \mathbf{I}$. 
Moreover, practitioners structure the output of the generative model in a residual form. Hence, the model does not directly generate an image: it predicts the noise $\boldsymbol{\epsilon}_t$ added to the image at each time step.
Then, it is possible to reconstruct the image $\mathbf{x}_0$ by subtracting $\boldsymbol{\epsilon}_t$ to the input $\mathbf{x}_t$. In this setup, the training objective becomes the following:
\begin{equation}
    L_t = \mathbb{E}_{t, \mathbf{x}_0, \boldsymbol{\epsilon}_t} 
    \Big[
        \|\boldsymbol{\epsilon}_t - \boldsymbol{\epsilon}_\theta(\sqrt{\bar{\alpha}_t}\mathbf{x}_0 + \sqrt{1 - \bar{\alpha}_t}\boldsymbol{\epsilon}_t, t)\|^2 
    \Big]
\end{equation}
where $L_t$ is the loss function evaluated at time step t. We leave interested readers refer to \cite{ho2020denoising,nichol2021improved} for the mathematical derivation and additional details.

\paragraph{Stable Diffusion.}
A widely adopted variant of DDMs is Latent Diffusion Models (LDMs). LDMs perform the diffusion process in the latent dimension rather than pixel space, saving training costs and ensuring faster inference.
In LDMs, an encoder $\mathcal{E}(\cdot)$ compresses an input image to a lower dimensional space $\mathbf{z}$, where the diffusion and denoising processes happen. 
Finally, we provide the generated sample as input to a decoder $\mathcal{D}(\cdot)$, trained to reconstruct images from the latent, $\tilde{\mathbf{x}} = \mathcal{D}(\mathbf{z})$. 

Stable Diffusion \cite{rombach2022high} is a popular LDM employing cross-attention layers to allow generic conditioning inputs such as text. 
It is possible to train the model using two regularisation terms: i) \textit{KL-reg}, which moves the learned latent toward a Standard Normal distribution, and ii) \textit{VQ-reg}, which uses a vector quantisation layer within the decoder (similar to VQGAN \cite{Esser_2021_CVPR}, but with the quantisation layer included in the decoder).
Stable Diffusion was trained on a large-scale dataset of natural images, showing impressive results in many applications. 
The model's pre-trained weights are available online,\footnote{https://huggingface.co/spaces/stabilityai/stable-diffusion} which makes it possible to use it as a foundation model for downstream tasks.

\paragraph{Learning New Concepts with Stable Diffusion.}\label{subsec:background_learning_new_concepts}
Training high-capacity generative models from scratch is challenging and requires high computational power and resources.
For this reason, in the recent literature, several approaches attempt to reuse large text-to-image diffusion models in different tasks, letting them learn new concepts using just a few examples and little computational power.

Low-Rank Adaptation of Large Language Models (LoRA) \cite{hu2021lora} is a training method that accelerates the training of large models while reducing memory usage. It freezes a pre-trained Stable Diffusion and injects trainable rank decomposition matrices into the model's transformer layers. 
These matrices are the only parameters trained during fine-tuning, reducing the number of trainable parameters for downstream tasks. After tuning the trainable weights on a dataset of new images, the model has learned the new concepts with little computational requirements.
Since the pre-trained Stable Diffusion already learned a powerful image prior, Textual Inversion \cite{gal2022image} proposes to learn new concepts by optimising the text embedding used as a text conditioning rather than the diffusion model itself. In particular, Textual Inversion captures a novel concept optimising the text embedding of a pseudo word while keeping the text-to-image model frozen. 
The optimisation proceeds until we find an optimal embedding vector that, used as a conditioning, lets the model generate samples of the new concept. 
DreamBooth \cite{ruiz2023dreambooth} is a technique that, given a small subset of images depicting a specific subject, synthesizes that
subject in diverse scenes, poses, views, and lighting conditions that do not appear in the reference images. 
This technique expands the language-vision dictionary of a pre-trained text-to-image model, fine-tuning it and binding a new unique identifier with what the user wants to generate. 

For its popularity and ability to generate high-quality images, we use DreamBooth to make our model capable of learning new concepts. Nevertheless, the model could learn new concepts using LoRA, Textual Inversion, or other approaches. We leave their investigation as future work.

\begin{figure}
    \centering
    \includegraphics[width=\columnwidth]{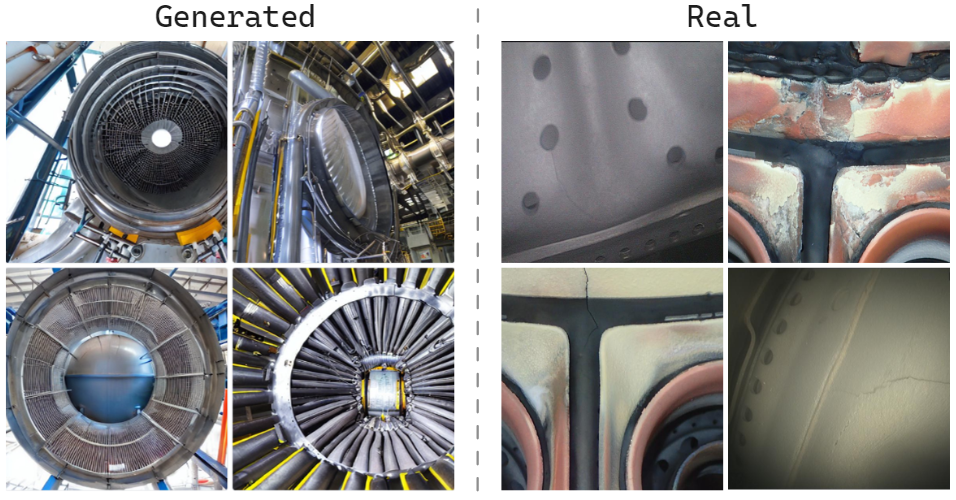}
    \caption{
        Generic prior vs real industrial images. The image prior learned by pre-trained foundation models, such as Stable Diffusion, do not match actual industrial data.
        On the \textbf{left}, we show four images generated by Stable Diffusion with the input prompt: {\fontfamily{pcr}\selectfont "A borescope image of the inside of a combustor chamber"}. On the \textbf{right}, we report four real industrial images, for comparison. We suggest to overcome this change in distribution as described in \Cref{subsec:learn_the_concept}.
    }
    \label{fig:prior}
\end{figure}

\section{Method}\label{sec:method}
\paragraph{Assumptions About the Data.}
Acquiring large-scale annotated datasets can be challenging, as industrial defects are rare and annotations are expensive. On the contrary, unlabelled images are usually easier to collect, e.g., as part of standard process monitoring activities.
In the rest of the paper, we assume the availability of a small amount of labelled data. The data cardinality can be small (in some of our experiments, we used just 120 annotated images). Furthermore, we suppose the availability of another subset of unlabelled images. 
More in detail, labelled data are images showing production defects manually annotated by an expert; unlabelled data are defect-free images acquired by standard monitoring activities.
To prevent any possible bias due to data leakage, we train our model using only the labelled data; we use the unlabelled images solely for the downstream data generation process.

\begin{figure}
    \centering
    \includegraphics[width=0.9\columnwidth]{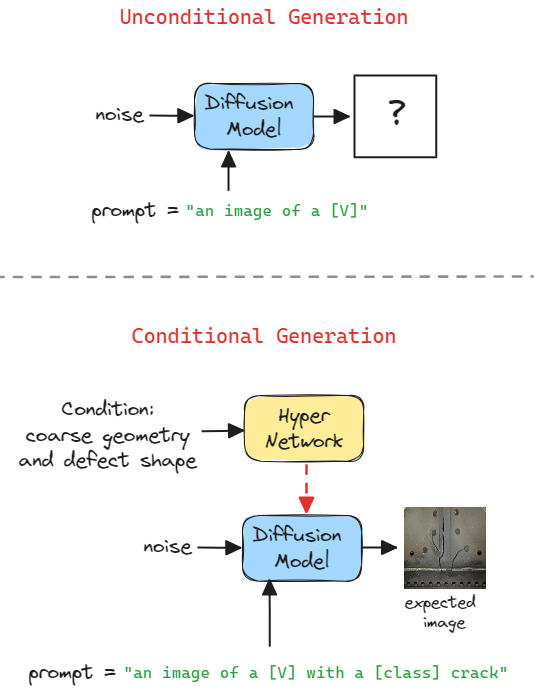}
    \caption{
        Conditioning pre-trained unconditional diffusion models with HyperNetworks. The HyperNetwork (\textbf{\color[HTML]{ffd43b} yellow} box) influences the image generator (\textbf{\color[HTML]{4dabf7}cyan} box) to output an image satisfying the input conditioning factors. As conditioning mechanism we use coarse geometrical drivers and a defect mask. Furthermore, we condition the diffusion model by slightly altering the input prompt from {\fontfamily{pcr}\selectfont "an image of a [V]"} to {\fontfamily{pcr}\selectfont "an image of a [V] with a [class] crack"}. For each condition, we replace {\fontfamily{pcr}\selectfont [class]} with the name of the category shown inside the defect mask, further highlighting crack-related information.
    }
    \label{fig:hypernet}
\end{figure}

\paragraph{Method Overview.}
Our approach exploits the image-prior learned by a foundation generative model. In particular, we use Stable Diffusion, whose weights come from training on a large-scale dataset of natural images.

With the right strategy, we can exploit Stable Diffusion's knowledge about images and their structure to learn to generate pictures of new concepts.
For us, the new concepts are industrial components whose pictures fall out of the training distribution seen by the model in precedence. 
After injecting the knowledge of the new concept, we force the generative process to adhere to geometrical constraints and produce images satisfying well-defined labels and conditions. 

We thus summarise the whole system as consisting of two main steps: i) \textit{learn the concept} and ii) \textit{learn the condition}. 
For both phases we use the small subset of labelled data, containing images and segmentation masks of the defects.
We provide a graphical overview of the two phases in \Cref{fig:graphical_abstract}, and detail them below.

\subsection{Phase 1: Learn the concept}\label{subsec:learn_the_concept}
In general, natural images used to train foundation models are not representative of data distributions encountered in industrial environments.
For example, \Cref{fig:prior} compares the ``idea" that Stable Diffusion has about the concept ``the inside of a gas combustor" vs real images. It is evident that directly using pre-trained image priors for generating industrial data is not feasible.
Thus, we must consider a preliminary step, where we adapt the foundation model to learn the new concept represented by the data of interest.

To inject the new concept into the model, we adopt DreamBooth \cite{ruiz2023dreambooth}, described in \Cref{subsec:background_learning_new_concepts}. With DreamBooth, we tune Stable Diffusion to produce realistic images while preserving prior knowledge about generic image appearance, such as lighting conditions, structural consistency, etc. 
To avoid catastrophic forgetting of the knowledge acquired by the pre-trained model, we regularise training on the new concepts with the Class-specific Prior Preservation Loss, which supervises the model with its own previously generated images.
Specifically, we use the ancestral sampler of the pre-trained DDM to generate data $\mathbf{x}_{pr} = \hat{\mathbf{x}}(\mathbf{z}_{t_1}, \mathbf{c}_{pr})$ from random initial noise $\mathbf{z}_{t_1}\sim\mathcal{N}(0, I)$ and conditioning vector $\mathbf{c}_{pr} \defeq \Gamma(f(\text{\fontfamily{pcr}\selectfont"a [class noun]"}))$. 
Then, we minimise:
\begin{equation}\label{eq:prior_pres1}
    \mathbb{E}_{\mathbf{x}, \mathbf{c}, \boldsymbol{\epsilon}, \boldsymbol{\epsilon}^\prime, t}\Big[ 
        \omega_t 
        \| \hat{\mathbf{x}}_\theta (\alpha_t \mathbf{x} + \sigma_t \boldsymbol{\epsilon}, \mathbf{c}) - \mathbf{x} \|^2_2 + \lambda L_{pr}
   \Big]
\end{equation}
\begin{equation}\label{eq:prior_pres2}
   L_{pr} = \omega_{t^\prime} \| \hat{\mathbf{x}}_\theta (\alpha_{t^\prime} \mathbf{x}_{pr} + \sigma_{t^\prime} \boldsymbol{\epsilon}{^\prime}, \mathbf{c}_{pr}) - \mathbf{x}_{pr} \|^2_2
\end{equation}
where $\alpha_t$, $\sigma_t$, and $\omega_t$ are terms controlling the noise diffusion schedule and sample quality at time $t \sim \mathcal{U}([0, 1])$, $\alpha_t \mathbf{x} + \sigma_t \boldsymbol{\epsilon}$ is the combination of a real image with the noise according to the diffusion schedule, $\mathbf{c}$ the text conditioning, $L_{pr}$ the prior-preservation term, and $\lambda$ a scalar weight.

To learn the new concept, we tune Stable Diffusion by assigning a new meaning to a rare token identifier [V]. To do so, we train the model on our data while providing the input prompt: {\fontfamily{pcr}\selectfont "an image of a [V]"}. As discussed in the next paragraphs, the second phase of our approach will slightly modify this textual prompt to include additional knowledge on the desired image characteristics. 

\subsection{Phase 2: Learn the condition}\label{subsec:learn_the_condition}
To make it possible to generate samples according to pixel-level requirements, we altered the behaviour of a pre-trained Stable Diffusion to make it accept additional inputs.
Similar to Zhang et al. \cite{zhang2023adding}, we control the synthesis process via HyperNetworks, which showed success in many image generation tasks \cite{ha2017hypernetworks,alaluf2022hyperstyle,dinh2022hyperinverter,ruiz2023hyperdreambooth}. As schematically represented in \Cref{fig:hypernet}, HyperNetworks \cite{ha2017hypernetworks} are small neural networks trained to influence the weights of a larger one based on input conditions. 
In our case, the network we want to influence is Stable Diffusion after learning the new concepts according to \Cref{subsec:learn_the_concept}. For the whole process, we keep this model frozen so that training preserves the starting weights of the model. 
On the contrary, we tune a trainable copy of Stable Diffusion that we use as a HyperNetwork to condition the frozen copy. 
Lastly, our input condition is a topological driver showing the coarse geometry of the image content and the defect class and shape. 
During this phase, the training objective becomes:
\begin{equation}\label{eq:learn_the_condition}
    \mathbb{E}_{\mathbf{x}, \mathbf{c}, \mathbf{c}^+, \boldsymbol{\epsilon}, t}\Big[ 
        \| \hat{\mathbf{x}}_\theta (\mathbf{z}_{t}, \mathbf{c}, \mathbf{c}^+) - \mathbf{x} \|^2_2
   \Big]
\end{equation}
where $\mathbf{z}_{t}=\alpha_t \mathbf{x} + \sigma_t \boldsymbol{\epsilon}$, while $\textbf{c}$ is the text prompt condition, and $\mathbf{c}^+$ is the task-specific condition. We describe the design of the topological drivers $\mathbf{c}^+$ below. 
For this phase, we use $\textbf{c} =~${\fontfamily{pcr}\selectfont "an image of a [V] with a [class] crack"}, where {\fontfamily{pcr}\selectfont [class]} is the category name of the crack we want the generated image to show.

\begin{figure}[t]
    \centering
    \includegraphics[width=0.9\columnwidth]{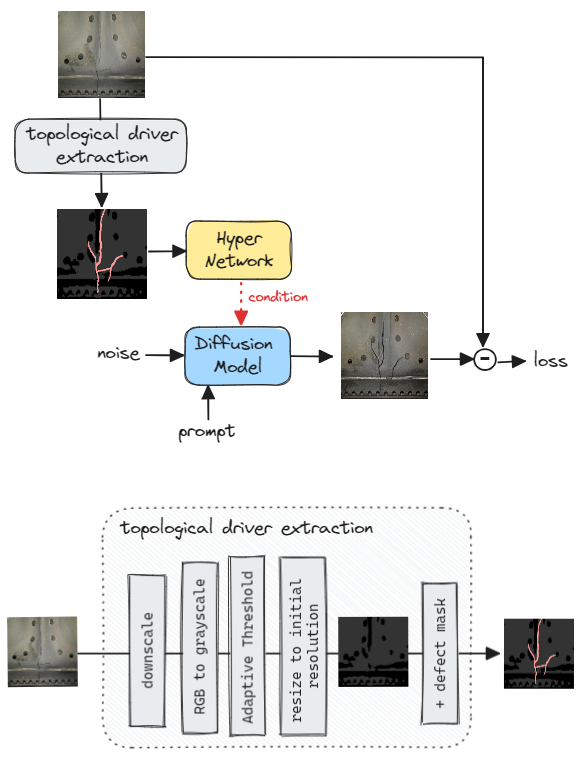}
    \caption{
    Model Overview.
    \textbf{Top}: after learning the new concept (\Cref{subsec:learn_the_condition}) a HyperNetwork (\textbf{\color[HTML]{ffd43b} yellow} box) uses an input topological driver to condition the image generation of a pre-trained diffusion model (\textbf{\color[HTML]{4dabf7}cyan} box). With the given condition, the model produces an image matching the topological driver and showing a defect corresponding to the defect mask inside the driver. 
    \textbf{Bottom}: extracting the topological driver from an input image. In the first part of the process, we do a lossy compression of the image colours. Then, we add the defect mask on the resulting image, ensuring that the driver carries out accurate pixel-level information of the defect topology and class. The resulting image shows coarse information about the image topology and a segmentation mask of the defect that the generative model must reproduce.
    }
    \label{fig:model_overview}
\end{figure}

\begin{figure}
    \centering
    \includegraphics[width=0.9\columnwidth]{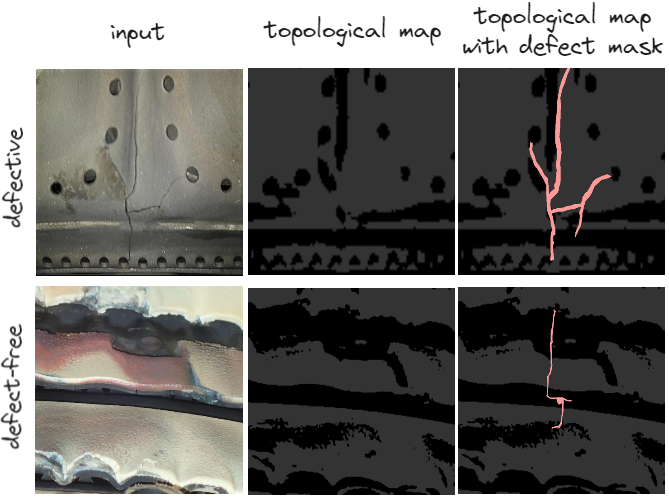}
    \caption{Topological drivers. 
    \textbf{Top row}: at \textit{training}, we extract the driver from defective images and add a real defect segmentation mask on top of it. 
    \textbf{Bottom row}: at \textit{inference}, we extract the driver from defect-free images and add an arbitrary segmentation mask on the image. Notice that simply looking at the final result, it is impossible to say if the topological map with the defect mask comes from a defective or a defect-free image. Hence, a model trained using only defective data will be biased to synthesise a defective image also for the input conditioning in the bottom row.}
    \label{fig:topological_driver}
\end{figure}

\paragraph{Building the Conditioning Image.}
While the designed topological driver well suits our use case, different datasets may need slightly different driver extraction procedures. Nevertheless, every driver extraction process should adhere to the same spirit: obtained drivers must provide i) coarse global constraints about the image appearance; and ii) strong constraints about the defect location and appearance. The first requirement ensures that the driver is flexible enough to let the model generate realistic alternative images sharing the same global aspect (e.g. with different colours or having some geometrical changes). The second constraint ensures we have the exact knowledge about the defect shape, class, and location, which is fundamental to having reliable annotations for the synthesised images.

To learn to generate defective images with conditioning constraints, we extract topological drivers from a small subset of defective images for which we have the segmentation masks of the defects. Then, we train the HyperNetwork to make the model reconstruct the defective image starting from the topological driver. 
The most important thing to notice is that we design topological drivers s.t. it is impossible to say if they come from defective or defect-free images.
Such a design ensures we can use images with no defects to drive generation at test time (see example in \Cref{fig:topological_driver}), thus making it possible to generate a wide variety of defective samples showing a topology that can be quite different from what was encountered during training.
As a result, we can obtain synthetic images with larger mode coverage than what would be attainable using only the annotated samples.

As depicted in \Cref{fig:model_overview}, the topological driver is an RGB image containing: coarse topological information about the image we want to generate and a defect mask painted on top of it. 
To extract the topological information, we first downscale the image to a lower resolution, then convert it to grayscale colours and apply adaptive thresholding. 
The resulting image is a binarised map showing the coarse topology of the input image at low resolution. After that, we resize the image back to the original size. We note that the pixels associated with a defect usually are only a tiny fraction of the whole image. 
As a result, the training process could encourage image reconstruction and ignore the defect completely, as they do not give a large contribution to the image reconstruction loss we are trying to minimise. 
To ensure the HyperNetwork forces the diffusion model to also focus on the defective pixels, we alter the binarised map to obtain an effect similar to an attention mechanism. 
First, we scale pixel values from the $0\div255$ range to the $0\div50$ range. Then, we inpaint an RGB-coloured defect mask on top of the image. Rescaling the binarised image to the $0\div50$ range ensures the defect mask is well visible and visually ``pops up" from the background. Lastly, we rescale the obtained images in the $0\div1$ range. 
We show examples of the obtained topological drivers in \Cref{fig:topological_driver}. Notice that the defect is well visible compared to the background. 

\subsection{Synthesis of Self-annotated Data}\label{subsec:synthesis_of_self_annotated_data}
Once learned the new concept and tuned the model to generate samples according to specific constraints, we can generate self-annotated data useful for downstream tasks.
In particular, we extract the topological drivers from a separate subset of defect-free images and pair them with arbitrary defect masks. 
The defect-free images come from standard monitoring activities and do not need annotations. The defect masks can be either i) randomly sampled from the masks available in the annotated dataset; ii) manually drawn by an expert, based on previous experience; or iii) could be generated by another neural network. 
To ensure good image-mask pairs, in our experiments, we build a small dataset using strategy (i) with small random shape perturbations and then (ii) to fix and improve the obtained masks. We leave exploring strategy (iii) as future work.

For a fixed topological driver and textual prompt, the diffusion model can generate an arbitrary number of synthetic images using different noisy inputs. For simplicity, we sample a maximum of 15 images per driver, which we deemed sufficient for our scope. We use the source defect masks as accurate segmentations of the synthetic images.

\section{Experiments}\label{sec:experiments}

\subsection{Data}
We conduct experiments on an in-house dataset named Turbine Boroscope Inspection (TBI). Thedataset contains RGB borescope pictures of gas turbines, compressors, and combustion chambers. 
Several images show cracks (i.e. defects) deriving from collisions or wearing processes, while others are defect-free. 
We show a few examples on the right column in \Cref{fig:prior}.
For the defective images, the dataset contains instance-level segmentation masks of each crack. Annotations are stored in the popular COCO format, that we handle using the \code{COCOHelper} library.\footnote{\url{https://github.com/AILAB-bh/cocohelper}}
The number of images in the dataset is 1470, out of which around 1300 have defects. The total number of annotated cracks is 2044, and their label is one of five classes, depending on their position and topology. 
For training, we only used 1170 annotated images. We left the remaining data for validating the experiments of instance segmentation models (we equally split the remaining data to build a validation and a test set, used to measure the performance of crack segmentors).

For both ``learn the concept" and ``learn the condition", we use only a fraction of the defective images, which we specify when describing the experiments. 
For both phases, we intentionally hold out defect-free samples, which we use only during inference, limiting the possibility of data leakage problems.
On the contrary, we always use all the defect-free images during synthesis. 
In particular, we first apply offline data augmentation, generating 10 alternative versions of the source image. Augmentations include geometrical and colour-wise transformations, such as cropping, zooming, roto-translation, noise addition, random contrast perturbations, etc. Then, we build a new dataset pairing each augmented sample with arbitrary defect masks, according to the strategies explained in \Cref{subsec:synthesis_of_self_annotated_data}. We name the resulting dataset SADF-TBI, where the prefix ``SADF" stands for ``Shuffled Annotations, Defect-Free".

We pre-process images by resizing them to the resolution of 512$\times$512 pixels. Then, we rescale pixel values in the 0$\div$1 range according to Stable Diffusion expected input dynamics. We also use standard data augmentation.

\begin{figure}
    \centering
    \includegraphics[width=0.85\columnwidth]{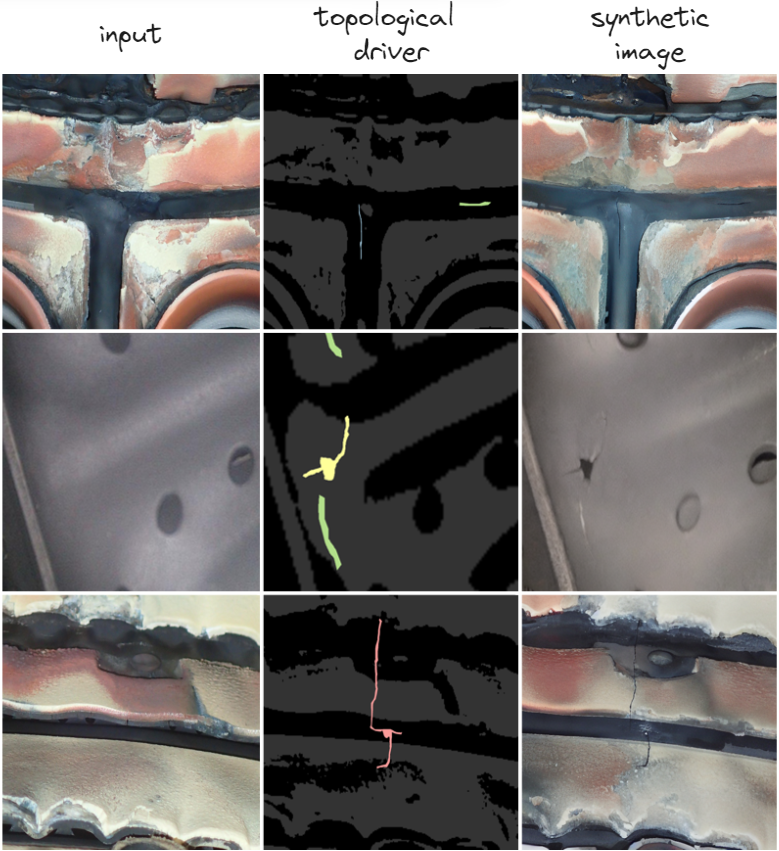}
    \caption{Example of synthetic data. 
    These samples have been generated according to the proposed approach using the input prompt: {\fontfamily{pcr}\selectfont "A borescope image of the inside of a combustor chamber with a [class(es)] crack"}. 
    As you can see by comparing the synthetic images with those generated in \Cref{fig:prior}, using our approach, Stable Diffusion learns to generate realistic data that were previously out of training distribution.
    Moreover, the generated pictures show cracks having geometry and position coherent with the input masks.
    Lastly, observe that the generated image colours may differ from the starting image, as this is not an inpainting process: on the contrary, the model generates the entire image conditioned only on the topological driver (see discussion in \Cref{subsec:our_vs_inpaint}).
    }
    \label{fig:generated_samples}
\end{figure}

\subsection{Main Results}
The primary objective of our study is to generate high-quality synthetic images for industrial use cases. As annotating data is expensive, the generated images should also contain defects in the exact locations indicated in the input topological driver. 
As depicted in \Cref{fig:generated_samples}, our method successfully synthesises realistic images. Notably, the global topology of the generated images follows the geometries in the input driver. Furthermore, they show defects whose appearance is coherent with the given defect masks.

We quantitatively evaluate the diversity and realism of the generated data using the Frechet Inception Distance (FID) \cite{heusel2017gans}. By comparing the distribution of synthetic images with that of the TBI data, we obtain a FID of 96.57.

Additionally, we investigate if we can use the self-annotated synthetic data on a real-industrial use case, consisting of training a model to perform instance segmentation of cracks. Specifically, we optimised Mask RCNN \cite{he2017mask} in three different settings: i) train on real data only; ii) train on synthetic data only; and iii) pre-train on synthetic, tune on real. Then, we measure performance on a held-out dataset and report results in \Cref{tab:mask_rcnn_1000}. As can be seen, using synthetic data for training consistently improves crack detection and segmentation for all the considered metrics. 

We also experimented with a more challenging set-up, using only 10\% of the dataset images (\Cref{tab:mask_rcnn_120}. Interestingly, we were not able to directly optimise Mask R-CNN on such a small dataset, as training quickly collapsed.\footnote{We experimented with several values of learning rate and batch size. In all the cases, the model collapsed to detect nothing or training diverged.} Instead, enlarging the dataset with our approach and then using both synthetic and real data led to good performance, close to using the entire dataset (IoU of 79.9\% vs 80.9\%).

\begin{table}\label{tab:mask_rcnn_1000}
    \centering
    \resizebox{\columnwidth}{!}{%
        \begin{tabular}{l|c|c|c|c}
               & Segm mAP ($\uparrow$)  &  BBox mAP ($\uparrow$)  &  IoU ($\uparrow$)  &  HD ($\downarrow$)  \\
            \midrule
            Synthetic Only   
                 & 0.06 & 0.12 & 80.6 & 0.29 \\
            Real Only   
                 & 0.12 & 0.27 & 80.7 & 0.19 \\
            Synthetic $+$ Real  
                 & \textbf{0.15} & \textbf{0.32} & \textbf{80.9} & \textbf{0.18} \\
            \bottomrule
        \end{tabular}
    }
    \caption{
        Instance Segmentation with synthetic data.  
        As can be seen, synthetic data allow to increase model performances according to all the considered metrics.
        We also provide results obtained by training our method only on synthetic data, showing that synthetic images well represent the data distribution and allow to train a good detector with no real data at all.
        Best results in \textbf{bold}.
    }
\end{table}

\begin{table}\label{tab:mask_rcnn_120}
    \centering
    \resizebox{\columnwidth}{!}{%
        \begin{tabular}{l|c|c|c|c}
               & Segm mAP ($\uparrow$)  &  BBox mAP ($\uparrow$)  &  IoU ($\uparrow$)  &  HD ($\downarrow$)  \\
            \midrule
            Synthetic Only   
                 & 0.02 & 0.04 & 75.85 & 0.42  \\
            Real Only   
                 & 0.00 & 0.00 & 0.00 & 1.00 \\
            Synthetic $+$ Real  
                 & \textbf{0.07} & \textbf{0.16} & \textbf{79.70} & \textbf{0.22}  \\
            \bottomrule
        \end{tabular}
    }
    \caption{
        Instance Segmentation with synthetic data.  
        Results refer to training using only \textbf{10\% of the annotated data}. 
        As can be seen, training high-capacity instance segmentation models, such as Mask RCNN, with such a small dataset is not feasible, and the model collapses. On the other hand, including a large amount of synthetic data stabilises training.
        Best results in \textbf{bold}.
    }
\end{table}

\subsection{Why not Inpainting?}\label{subsec:our_vs_inpaint}
As mentioned in \Cref{sec:related_work}, inpainting methods may not fit our use case since the generated images would not cover the diversity of industrial data. 
To highlight this further, we conducted a simple experiment, adapting the model to do inpainting. 
After the ``learn the concept" phase, we trained the model to generate data directly conditioned on the source RGB image with its defect masks overlaid. 
By pasting the defect mask into the image, we removed local information about the crack's appearance, leaving only geometrical attributes. Notice that such a conditioning keeps the background information while the coarse topological driver in \Cref{subsec:learn_the_condition} loses most of it.
To output realistic pictures, the model must learn to inpaint defects in the mask location and harmonise it with the surrounding pixels.

We trained the model on the TBI dataset and generated synthetic images using SADF-TBI (i.e. defect-free images paired with defect masks). 
To measure generated data variety, we kept the conditioning image fixed and sampled multiple times from Gaussian noise. Then, we sampled $k$ different samples for $m$ conditioning images, where $m=k=10$, for a total of 100 synthetic images.
We would expect high variety of the generated data if: i) the average L2 Distance between generated samples is high; ii) the Mutual Information is low; and iii) the FID between the generated images and the real dataset is low.
As shown in \Cref{tab:inpaint}, the inpainting method is inferior to our approach. 

\begin{table}
    \caption{
        Comparison with inpainting.
        We measure the generated data variety by keeping the same input conditions and sampling multiple times from a Normal distribution. As can be seen, our method achieves better mode coverage compared to simple inpainting. 
        Best results in \textbf{bold}.
        }
    \label{tab:inpaint}
    \centering
    \resizebox{\columnwidth}{!}{%
        \begin{tabular}{l|c|c|c}
             & L2 Distance ($\uparrow$) & Mutual Information ($\downarrow$) & FID ($\downarrow$) \\
            \midrule
            Inpainting 
                & 64.17 & 1.36 & 203.83 \\
            Ours 
                & \textbf{91.36} & \textbf{0.70} & \textbf{181.15} \\
            \bottomrule
        \end{tabular}
    }
\end{table}

\subsection{Ablation Study}
We investigate learning the concept before learning to condition. We expect that learning the new concept first is especially useful with small datasets. Thus, we analyse the FID obtained by the model using only 10\% of the TBI dataset. 
As shown in \Cref{tab:ablation}, introducing the new concept in a preliminary phase helps to better understand the data distribution, useful for the subsequent generation step.

\begin{table}
    \caption{
        Ablation study. 
        Learning the new concept before learning to condition improves performance (\#1 vs \#2). For comparison, we also report our method when training on the whole training set, as a lower-bound for the FID (\#3).
    }
    \label{tab:ablation}
    
    \centering
    \resizebox{0.9\columnwidth}{!}{%
        \begin{tabular}{l|c|c|c|c}
            & Learn the  
            & Learn the  
            & \multirow{2}{*}{Training data}
            & \multirow{2}{*}{FID ($\downarrow$)} \\
            & Concept
            & Condition
            & & \\            
            \midrule
            \#1 &  & $\checkmark$
                & 10\%  & 106.17 \\
            \#2 & $\checkmark$ & $\checkmark$
                & 10\% & 102.79 \\
            \midrule
            \#3 & $\checkmark$ & $\checkmark$
                & 100\% & 96.57 \\
            \bottomrule
        \end{tabular}
    }
\end{table}


\section{Conclusion}\label{sec:conclusion}
We introduce an approach for enlarging small industrial datasets through general-purpose foundation generative models. In particular, we use a pre-trained text-to-image diffusion model, Stable Diffusion, that we adapt to generate images that contain novel concepts and are self-annotated for downstream tasks. We base our approach on two distinct steps: i) \textit{learn the concept} and ii) \textit{learn the condition}. Our experiments show that the generated self-annotated data have high quality and can be used to train instance segmentation models for industrial applications. 

\bibliographystyle{splncs04}
\bibliography{references}
\end{document}